\documentclass[conference]{IEEEtran}
\usepackage[utf8]{inputenc}
\IEEEoverridecommandlockouts
\usepackage{cite}
\usepackage{amsmath,amssymb,amsfonts}
\usepackage{algorithmic}
\usepackage{graphicx}
\usepackage{url}
\usepackage[breaklinks=true]{hyperref}
\usepackage{xurl}
\usepackage{textcomp}
\usepackage{xcolor}
\usepackage{url}
\usepackage{booktabs}
\usepackage{multirow}
\usepackage{enumitem}
\usepackage{adjustbox}

\newcommand{\gptiii}{{\fontfamily{lmtt}\selectfont GPT-3.5 Turbo}}
\newcommand{\gptivo}{{\fontfamily{lmtt}\selectfont GPT-4o mini}}
\newcommand{\gemini}{{\fontfamily{lmtt}\selectfont Gemini 1.5 Pro}}

\newcommand{\eg}{\textit{e}.\textit{g}., }

\def\BibTeX{{\rm B\kern-.05em{\sc i\kern-.025em b}\kern-.08em
    T\kern-.1667em\lower.7ex\hbox{E}\kern-.125emX}}
\begin{document}

\title{Assessing Historical Structural Oppression Worldwide via Rule-Guided Prompting of Large Language Models
}

\makeatletter
\newcommand{\linebreakand}{%
  \end{@IEEEauthorhalign}
  \hfill\mbox{}\par
  \mbox{}\hfill\begin{@IEEEauthorhalign}
}
\makeatother

\author{
\IEEEauthorblockN{Sreejato Chatterjee}
\IEEEauthorblockA{\textit{Goergen Institute for Data Science} \\
\textit{University of Rochester}\\
Rochester, NY, USA \\
schatte9@u.rochester.edu}
\and
\IEEEauthorblockN{Linh Tran}
\IEEEauthorblockA{\textit{Department of Computer Science} \\
\textit{University of Rochester}\\
Rochester, NY, USA \\
ltran18@ur.rochester.edu}
\and
\IEEEauthorblockN{Quoc Duy Nguyen}
\IEEEauthorblockA{\textit{Department of Economics} \\
\textit{University of Rochester}\\
Rochester, NY, USA \\
qnguy12@u.rochester.edu}
\linebreakand
\IEEEauthorblockN{Roni Kirson}
\IEEEauthorblockA{\textit{College of Arts and Sciences} \\
\textit{University of Rochester}\\
Rochester, NY, USA \\
rkirson@u.rochester.edu}
\and
\IEEEauthorblockN{Drue Hamlin}
\IEEEauthorblockA{\textit{Department of Sociology and Anthropology} \\
\textit{Rochester Institute of Technology}\\
Rochester, NY, USA \\
dah2151@g.rit.edu}
\and
\IEEEauthorblockN{Harvest Aquino}
\IEEEauthorblockA{\textit{Department of Public Health} \\
\textit{University of Rochester}\\
Rochester, NY, USA \\
haquino2@u.rochester.edu}
\linebreakand
\IEEEauthorblockN{Hanjia Lyu}
\IEEEauthorblockA{\textit{Department of Computer Science} \\
\textit{University of Rochester}\\
Rochester, NY, USA \\
hlyu5@ur.rochester.edu}
\and
\IEEEauthorblockN{Jiebo Luo}
\IEEEauthorblockA{\textit{Department of Computer Science} \\
\textit{University of Rochester}\\
Rochester, NY, USA \\
jluo@cs.rochester.edu}
\and
\IEEEauthorblockN{Timothy Dye}
\IEEEauthorblockA{\textit{Department of Obstetrics and Gynecology} \\
\textit{University of Rochester School of Medicine}\\
Rochester, NY, USA \\
tim\_dye@urmc.rochester.edu}
}

\maketitle

\begin{abstract}
Traditional efforts to measure historical structural oppression struggle with cross-national validity due to the unique, locally specified histories of exclusion, colonization, and social status in each country, and often have relied on structured indices that privilege material resources while overlooking lived, identity-based exclusion. 
We introduce a novel framework for oppression measurement that leverages Large Language Models (LLMs) to generate context-sensitive scores of lived historical disadvantage across diverse geopolitical settings. Using unstructured self-identified ethnicity utterances from a multilingual COVID-19 global study, we design rule-guided prompting strategies that encourage models to produce interpretable, theoretically grounded estimations of oppression. We systematically evaluate these strategies across multiple state-of-the-art LLMs. Our results demonstrate that LLMs, when guided by explicit rules, can capture nuanced forms of identity-based historical oppression within nations. This approach provides a complementary measurement tool that highlights dimensions of systemic exclusion, offering a scalable, cross-cultural lens for understanding how oppression manifests in data-driven research and public health contexts. To support reproducible evaluation, we release an open-sourced benchmark dataset for assessing LLMs on oppression measurement (\url{https://github.com/chattergpt/HSO-Bench}).
\end{abstract}


\section{Introduction}\label{sec:intro}

The study of racial and ethnic inequality remains central to sociological research, with extensive research documenting how structural oppression is reproduced in historical and contemporary contexts~\cite{braveman2021, bonillasilva1997, Saperstein2013}. Oppression can be understood as a social hierarchy in which some groups subject other groups to lower status and to systemic exclusion, dehumanization, and disadvantage. In public health and sociology, this oppression is closely aligned with definitions of systemic and structural racism, which describe racism as deeply embedded in laws, policies, institutional practices, and social norms that sustain widespread inequities, violence, and disadvantage over time~\cite{braveman2021}. Foundational works have demonstrated how ethnic and national hierarchies shape access to power, life opportunities, autonomy, and sovereignty, for example, primarily through institutionalized mechanisms such as legal structures, educational systems, and healthcare access, among others~\cite{bonillasilva1997}. These dynamics vary substantially across geopolitical regions, however, shaped by localized histories of colonization, racial and ethnic discrimination, and national classification systems ~\cite{brubaker2006,hall2000}. As such, understanding racial and ethnic oppression defies cross-national analyses due to their somewhat idiosyncratic and varying origins, requiring methodologies that are sensitive to national contexts and histories. 
Traditional frameworks often rely on demographic indicators such as income, housing, and education, and perhaps standardized categories of race and/or ethnicity specific to that country, but can hardly capture the full spectrum of structural marginalization experienced by ethnic and racial groups across countries~\cite{allik2020}. Additionally, oppression is a multidimensional phenomenon shaped not only by material violence but also by systemic exclusion from institutions, denial of access, fragility of social networks, and historically maintained systems of classification. Crucially, these processes are geographically and culturally contingent, varying across national boundaries and historical contexts~\cite{allik2020, basu2020, wang2021}. Achieving health equity requires measures that go beyond material deprivation and account for dimensions such as structural racism, historical injustice, and cultural exclusion, which can be underrepresented in current indices~\cite{braveman2021}. Capturing these dimensions is especially challenging due to the localized nature of structural oppression, as each country has its own distinct history of exclusion, marginalization, and racial classification systems. A group that is marginalized in one society may be dominant in another, making it impossible to apply a single, global framework without oversimplifying these dynamics.

Efforts to measure race and ethnicity in health and social research often rely on standardized categories imposed by government and statistical agencies, such as the Office of Management and Budget's (OMB) minimum standards for race and ethnicity reporting in the United States (OMB15)~\cite{OMB1997, OMB2016} or census classifications in other countries~\cite{Morning2008}. While these systems provide structured categories that enable comparability within national datasets, they frequently fail to capture how individuals actually identify~\cite{Saperstein2013}. Respondents are forced to choose categories that do not reflect their lived experience, and large umbrella groups such as ``Asian" and ``Hispanic" obscure substantial heterogeneity in histories, cultures, and health outcomes across subgroups. For example, aggregating South Asian, East Asian, and Southeast Asian populations under a single ``Asian" label masks wide disparities in disease risk, socioeconomic position and discrimination patterns~\cite{Islam2010}. Similarly, treating ``Hispanic/Latino" as a homogeneous category overlooks differences across Mexican, Puerto Rican, Cuban, and Central/South American populations in health outcomes and access to care~\cite{Lara2005}. These limitations highlight the core conflict between comparability and accuracy, as structured taxonomies facilitate statistical analysis but flatten identity-based variation, whereas free-text self-identification allows people to describe themselves in context-specific and intersectional ways but resists conventional quantitative analysis. These challenges underscore the need for new methodological approaches that can handle the flexibility of self-identified, unstructured identity data, motivating exploration of Large Language Models (LLMs) as tools for interpreting free-text inputs at scale.

Large Language Models (LLMs) have emerged as tools for use in many domains, especially social sciences~\cite{lyu2025gpt,mou2024unifying} and public health~\cite{jo2023understanding} to support tasks such as free-text extraction and classification, with high performance across a range of tasks when applied at scale~\cite{guo2024llmhealth}. However, among more sociological tasks, LLMs have presented their own issues and biases~\cite{qi2025representation}, including reproduction of racial and ethnic stereotypes and underrepresentation of structurally marginalized groups~\cite{sheng2021bias}.

Our work seeks to address the limitations of traditional deprivation indices by introducing a complement using free-text descriptions of self-identified ethnic group and country of residence as inputs. By prompting models to reason about local histories, systemic exclusion, and institutional access, while grounding them in sociological theories, we generate synthetic oppression scores that can be meaningfully compared across contexts through consistent prompt structures and scoring criteria. Although LLMs have limitations, their capacity to synthesize large-scale textual knowledge across languages makes them well-suited to address the challenge of modeling historical structural oppression across different countries through incorporating localized histories and national context without having to manually encode each country’s unique sociopolitical structure, which, in the absence of emic experts from within each country, is logistically impossible for most efforts. 
Through this approach, we aim to provide a scalable and interpretable framework for measuring systemic inequality across diverse populations.
To summarize, our contributions are threefold:
\begin{itemize}[leftmargin=*]
    \item We develop a bottom-up schema for historical oppression measurement derived from free-text responses of self-identified ethnicity and country of residence across the world. This provides a complement to existing indices of social disadvantage that emphasize material conditions.
    \item We scale this measurement using large language models, incorporating a rule-guided prompting module to enhance validity and reliability. Our method achieves a Pearson correlation coefficient of 0.852 with human expert annotations, demonstrating strong alignment.
    \item We release the dataset and benchmark results for measuring identity-based oppression, establishing a new task and resource for evaluating models on capturing lived, identity-based exclusion.
\end{itemize}

\section{Related Work}\label{sec:related_work}
\subsection{Quantifying Systemic Oppression Across Contexts}
Currently, most efforts to quantify social disadvantage, such as the Index of Multiple Deprivation (IMD) in the UK~\cite{imd2019} or regional deprivation indices in India~\cite{basu2020} and Brazil~\cite{allik2020brazdep}, have relied on composite measures built from structured datasets, including census records, administrative data, and geospatial indicators~\cite{wang2021, maroko2016, basu2020}, rather than flexible, open-ended utterances from people themselves. These tools, while valuable for resource allocation and public health monitoring, often suffer from three limitations. First, they privilege domains such as income and employment, often treating these metrics as identity-neutral and overlooking how race, ethnicity and structural power relations shape access to resources, thereby underrepresenting identity and power-based oppression dynamics. Second, they are often limited to national or subnational contexts and struggle with generalizability across borders. Third, their dependence on available administrative data often excludes experiences of discrimination that are not explicitly measured or encoded in existing survey methods~\cite{wang2021, maroko2016, basu2020}.

In addition to these limitations, standardized racial and ethnic categories used in census and administrative datasets are themselves politically derived and often carry inherent bias. For example, Native American respondents in the U.S. are often denied the opportunity to identify by tribal affiliation, despite the fact that tribal membership is central to accessing resources, legal status, and cultural identity~\cite{Snipp2003}. These classifiers serve administrative or political purposes rather than reflecting the lived experiences of those being categorized~\cite{Morning2008}. In many cases, the very populations that experience systemic oppression are either excluded from official classifications altogether or absorbed under categories designed for bureaucratic convenience. By contrast, unstructured self-identification enables individuals to articulate their identities in their own terms, thereby democratizing ethnic categorization and revealing nuanced intersectional positions that structured indices overlook. This flexibility is particularly important for capturing marginalized, diasporic populations or mixed-identity populations whose self-descriptions may not neatly map onto standardized census categories.

Developing oppression measurements requires careful selection of indicators grounded in theory, followed by standardization, weighting, and validation across contexts~\cite{allik2020}. However, there is no consensus on how to assign relevant domains in global or multi-ethnic contexts. Factor analysis techniques like Principal Component Analysis (PCA) and its geographically weighted variant (GWPCA) are often used to derive scores, but these approaches struggle with replicability over time. Additionally, while validation is commonly performed against health outcomes or existing indices, such efforts rarely disentangle the specific role of identity-based oppression from broader measures of material disadvantage~\cite{basu2020, wang2021}. 

These limitations reflect broader concerns in the biomedical and social science literature regarding the misuse and oversimplification of race and ethnicity in research, and the denial of autonomy in self-identifying one’s ethnicity. Race and ethnicity are frequently treated as static variables without sufficient consideration of their sociopolitical construction, contextual meanings, or structural forces that shape differential health outcomes. The National Academies of Sciences, Engineering, and Medicine have called for researchers to adopt principled, theory-driven definitions of race and ethnicity, disclose data provenance, and critically assess the rationale and limitations behind their use~\cite{powe2025race}. In alignment with these recommendations, our study conceptualizes oppression as an outcome of institutional power and historical injustice rather than simply a demographic descriptor.

\subsection{LLMs and Sociological Reasoning}
An emerging area of application for LLMs is survey research, where they show promise for generating synthetic responses, clarifying question-wording, and mitigating response bias by simulating varied interpretations of survey items~\cite{jansen2023survey,zhang2025socioverse}. These applications highlight how LLMs can be integrated into existing workflows to improve data quality and enhance interpretability with methodological care and rigor.

However, few studies have evaluated whether LLMs can produce valid, replicable judgments on concepts such as discrimination, oppression, or privilege, especially when grounded in sociological theory. 
Recent evaluations have primarily focused on fairness in text generation, with limited attention to how LLMs approximate sociological indices~\cite{gebru2021datasheets,srivastava2022beyond,liang2022holistic,qi2025representation}. Additionally, existing tools to quantify systemic oppression struggle with scalability and cross-national generalization, often relying on expert-coded taxonomies or language-specific analyses~\cite{allik2020, basu2020, wang2021}, or inappropriate adoption of one country’s predominant categories in another, which can effectively export one country’s racist construction of ethnicity to another country. This limitation underscores the need for flexible approaches that are adaptable across cultural and linguistic concepts, motivating our use of instruction-tuned LLMs guided by sociologically-informed prompts and scoring rubrics.

\section{Methods}\label{sec:method}
In Section~\ref{sec:task_formulation}, we describe how we construct a schema for identity-based oppression scoring. In Section~\ref{sec:llm_framework}, we present our LLM-based method for applying this scoring at scale.

\subsection{Schema Construction}\label{sec:task_formulation}
We adopt a \textbf{bottom-up approach} to construct a schema for identity-based historical oppression classification. We adopt a bottom-up approach to construct a schema for identity-based historical oppression classification. Our starting point is a secondary analysis of a multilingual global survey on COVID-19 non-medical COVID-related impact~\cite{dye2021covid}, where participants were asked to answer the open-ended question \textit{“How would you describe your ethnic background?”} (other language variants are provided in Appendix~\ref{appendix_sec:survey_question}) under the Demographic Questions section of the survey to self-identify their ethnic group. Participants were also asked to choose their country of residence from the standardized list of countries provided by International Standard Organization 3166 Country Codes, with an option for self-description if their country was not provided~\cite{iso_obp}. This is a decision that prioritizes each person’s current social context – regardless of the meaning of their identity in their birth country (if different) – which can change dramatically in changing from one country of residence to another. The resulting dataset contains a wide range of free-text responses across multiple languages (\eg English, Spanish, Italian, Mandarin, Arabic, ASL, French, Hindi), offering a diverse window into global identity categories. Unlike traditional census methods for identifying race or ethnicity, these free-text descriptions allow study participants to identify themselves and their own identities rather than choosing from pre-defined categories.

To capture a broad spectrum of geopolitical and sociocultural contexts, we sample 334 responses from 10 countries: the United States (n = 49), Canada (n = 43), Australia (n = 50), Brazil (n = 50), Algeria (n = 35), Palestine (n = 14), Puerto Rico (n = 49), Finland (n = 11), Madagascar (n = 12), and Sweden (n = 21). These countries are selected to maximize regional diversity (North America, Latin America, Europe, Africa, the Middle East, and Oceania) and to include both politically stable and conflict-affected settings, and areas of complex migrations that may result in high cultural heterogeneity.

We review and cluster these responses, guided by critical theoretical frameworks including cumulative disadvantage~\cite{dannefer2003cumulative}, racial formation~\cite{powe2025race}, historical trauma~\cite{braveheart1998american}, and critical race theory~\cite{bonillasilva1997, pierson2004politics}. From this analysis, we develop a \textbf{five-level ordinal classification} that conceptualizes oppression as a group’s institutional positioning, historical marginalization, and exposure to structural violence. The scale is theory-driven, as we do not assume equal intervals between categories and is not empirically calibrated via qualitative outcome data. Rather, it is derived from historical and sociopolitical context, reflecting increasing degrees of systemic exclusion and marginalization. The schema captures a comprehensive continuum and accommodates all observed responses:

\begin{enumerate}[leftmargin=*]
    \item \textbf{Little to No Oppression} – These groups constitute the most privileged in their society. They have not experienced group systemic oppression or violence on the basis of their identity, and they experience no excess social and health risk based on their ethnicity. They were historically aligned with or integrated into the dominant societal institutions, laws, or culture of their society and were able to easily assimilate into it. Over time, they have maintained extensive access to societal resources, opportunities, and benefits and experience the most institutional privilege. They are highly unlikely to encounter systemic barriers, discrimination, violence, or other oppression based on their race or ethnicity, and they do not encounter bias, stereotyping, or tokenism on the basis of their ethnic and/or racial identity.

    \item \textbf{Low Oppression} - These groups constitute the second most privileged in their society. They have experienced minimal systemic oppression or violence on the basis of their identity and have minimal social risk factors. They are allowed to assimilate with relatively few obstacles into the dominant institutions, laws, or culture of their society. Over time, they have maintained broad access to societal resources, opportunities, and benefits, and have experienced institutional privilege with some exceptions. They are moderately unlikely to encounter systemic barriers, discrimination, violence, or other oppression based on their race or ethnicity, but their groups may have encountered occasional bias, stereotyping, or tokenism on the basis of their ethnic and/or racial identity.

    \item \textbf{Moderate Oppression} - These groups constitute the middle group in terms of privilege versus oppression in their society. They have experienced moderate systemic oppression or violence on the basis of their identity, and have moderate social risk factors. Their groups have not easily been allowed to assimilate into the dominant institutions, laws, or culture of their society. Over time, they have maintained some access to societal resources, opportunities, and benefits and have experienced some institutional privileges. They have maintained some but incomplete access to institutional power, and face recurring bias, stereotyping, or discrimination. They are often visible in some domains but underrepresented or stereotyped in others.

    \item \textbf{High Oppression} - These groups have experienced longstanding, institutionalized exclusion, including colonization, forced assimilation, segregation, or legal discrimination. While they may have been partially included in legal or social systems, they were treated as subordinate populations with restricted rights, power, or recognition. Today, these groups often continue to face persistent systemic disadvantage and under-representation in governance, business, science, and other domains, with limited progress in inclusion or representation.

    \item \textbf{Severe Oppression} - These groups constitute the most oppressed group in their society. They have experienced severe levels of systemic oppression or violence on the basis of their identity and have historically experienced severe social risk factors. They have been, at least historically, actively excluded from the dominant institutions, laws, or culture of their society. Over time, they have experienced little access to societal resources, opportunities, and benefits, and have not as a group experienced frequent institutional privilege. They are the most likely in their societies to continue to encounter systemic barriers, discrimination, violence, severe health and social risk, or other oppression based on their race or ethnicity, and they very frequently encounter bias, stereotyping, or tokenism on the basis of their ethnic and/or racial identity.
\end{enumerate}

We then apply this schema to the 334 sampled responses, annotating each group’s oppression level based on their self-identified ethnicity and country of residence. Annotations are conducted by student researchers trained in sociological methods, with backgrounds in public health, political science, and anthropology. Each annotation is informed by external sources—including academic publications, historical accounts, and policy documents—and documented with detailed rationale and citations to ensure transparency and replicability. 

For example, Black Americans and Ashkenazi Jews are assigned higher historical oppression scores due to histories of slavery, antisemitism, and continuing structural violence. In contrast, groups such as Europeans or English people in the United States, Canada, and Australia receive lower historical oppression scores, reflecting historical dominance and institutional privilege. Although each group is ultimately assigned a single classification, the annotation process involves collaborative discussion, source-sharing, and consensus-building to ensure accountability.

\subsection{LLM-Based Oppression Annotation}\label{sec:llm_framework}
While effective human expert annotation provides the most reliable basis to assess oppression levels, it is inherently time-consuming, costly, potentially biased or incompletely-informed, and difficult to extend across diverse cultural and linguistic contexts. To address these limitations, we design an LLM-based method that approximates human annotation at scale. LLMs offer strong capabilities in reasoning, contextual understanding, and cross-linguistic generalization, making them particularly well-suited for interpreting free-text identity data in globally diverse surveys.

\subsubsection{Prompt Template and Components}\label{sec:prompt_template}
We construct a standardized prompt template with five components: system role, identity statement, task instruction, oppression schema, and required output format.  

\begin{itemize}[leftmargin=*]
    \item \textbf{System role.} The model is primed as a domain expert:  
“\emph{You are a knowledgeable cultural sociologist.}”
\item \textbf{Identity statement.} Each instance specifies the respondent’s country and self-identified group:  
“\emph{I am a person in \texttt{\{country\}}, and here is what I identified as: \texttt{\{identity\}}.}”
\item \textbf{Instruction.} The directive anchors outputs to the rubric:  
“\emph{Using these numerical categories as a scale, give a number and explanation that best describes my situation. Your response must be based on the given definitions. Do not infer additional factors outside the definitions provided.}”
\item \textbf{Oppression schema.} The rubric for levels 1–5 is attached.  
\item \textbf{Required output format.} A fixed, parseable structure is enforced:  
“\emph{Rating: \textless 1–5 \textgreater; Explanation: \textless brief explanation based on the context\textgreater.}”
\end{itemize}

Prompts are assembled in the following order: (1) system role $\rightarrow$ (2) identity statement $\rightarrow$ (3) instruction $\rightarrow$ (4) oppression schema $\rightarrow$ (5) required output format. This is illustrated in Figure~\ref{fig:promptchart}.

\subsubsection{Rule-Guided Prompt Module}\label{sec:rule_guided}
Preliminary experiments reveal that naive prompting often produces inconsistent or inaccurate results. Models tend to over-rely on recent events, general stereotypes, or globally marginalized assumptions that ignore national context. To address this, we introduce a \textbf{rule-guided prompt module}: a set of sociological instructions embedded in the prompt that constrain model reasoning and enforce context-aware, historically grounded scoring.  
The rules serve two purposes: (i) to ensure consistency with established frameworks of systemic oppression, and (ii) to enhance comparability across countries and languages.

\begin{enumerate}[leftmargin=*]
    \item This classification must be based solely on historical and systemic factors of oppression. Do not consider cultural contributions, economic success, or individual achievements when assigning a category.
    
    \item Do not assume that globally marginalized identities (e.g., Asian, Jewish, Latino) experience systemic oppression in the same way across all societies. Your classification must be based strictly on the historical and systemic role of that identity group within \texttt{\{country\}}.
    
    \item When someone identifies using a national label (e.g., ``Canadian,'' ``Brazilian,'' ``American''), assume they are referring to the dominant racial or ethnic group in that country, unless the label includes an additional modifier that indicates a minority or marginalized population.
    
    \item When someone identifies as having both privileged and historically oppressed ancestries, lean toward the rating that reflects the marginalized component, especially if the society has historically assigned group membership or social treatment based on that marginalized identity. Many societies treat such individuals as non-members of the dominant group, regardless of partial privileged heritage.
    
    \item If no evidence exists of structural disadvantage, assign a low score. Do not infer oppression based on general trends, recent events, or social stereotypes not grounded in the long-term history of systemic oppression in \texttt{\{country\}}.
    
    \item Only assign a 4 or higher if the group has faced long-term, institutionalized exclusion across multiple major domains (e.g., housing, education, voting, etc.), with limited inclusion efforts over time. Consider category 3 if the group has experienced discrimination, stereotyping, or underrepresentation, but has maintained meaningful access to education, employment, and civic institutions.
\end{enumerate}

These rules provide a principled scaffold for LLM reasoning, ensuring alignment with historical and sociological frameworks. For instance, Rule 2 enables differentiation of ``Hispanic” respondents in Brazil versus the United States, while Rule 4 acknowledges the impact of the historically pervasive ``one-drop rule" to correctly classify mixed-ancestry groups. The ``one-drop" was a legal and social doctrine in the United States that classified any individual with a trace of African ancestry as Black, and was codified in state statutes and upheld as a tool to enforce rigid racial boundaries during the Jim Crow Era~\cite{Khanna2010}. We reference the rule only as a historical example of ancestry-based exclusion, \textbf{\textit{not}} as an endorsement. While the one-drop rule is specific to the United States, similar logics of exclusion based on partial marginalized ancestry exist in other countries, albeit with different historical foundations. For example, in Brazil and other Latin American countries, individuals with mixed African ancestry may be socially classified in less marginalized categories depending on skin color or phenotype, yet still encounter discrimination tied to their ancestry~\cite{Telles2014}. Additionally, Rule 3 ensures consistency when national identifiers are used, defaulting to dominant groups unless modifiers indicate otherwise.

\section{Experiments}\label{sec:exp}

\begin{figure*}[htbp] % The starred version for spanning two columns
    \centering
    \includegraphics[width=\textwidth]{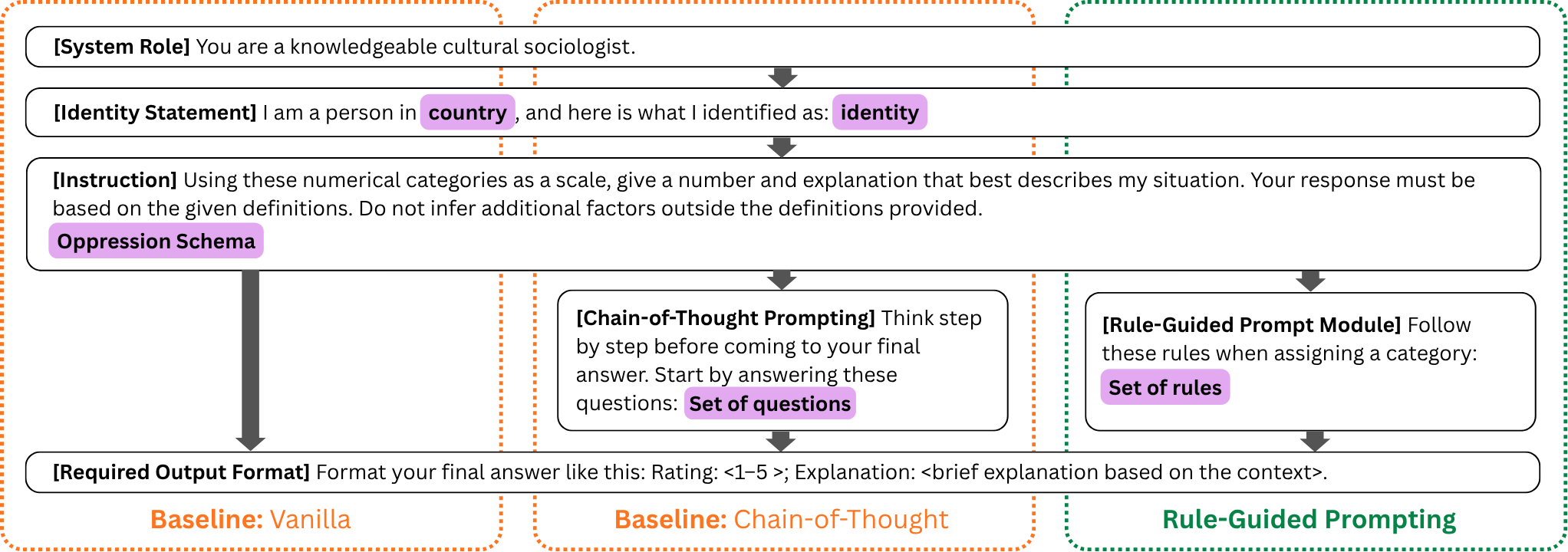} % Adjust width as needed
    \caption{Illustration of the prompting setup used for our experiment, comparing  Rule-Guided Prompting (RG) with two baseline methods: Vanilla and Chain-of-Thought (CoT). Each prompt is constructed sequentially as follows: (1) system role $\rightarrow$ (2) identity statement $\rightarrow$ (3) instruction $\rightarrow$ (4) oppression schema $\rightarrow$ (5) optional CoT or RG instruction $\rightarrow$ (6) required output format. The complete set of rules for the RG method is provided in Section~\ref{sec:rule_guided}, and the set of guiding questions for the CoT method is detailed in Section~\ref{sec:baseline}.
    }
    \label{fig:promptchart}
\end{figure*}

\subsection{Experiment Setup}\label{sec:exp_setup}
We evaluate our method on the task of identity-based oppression scoring using the 334 sampled responses introduced in Section~\ref{sec:task_formulation}. Each response is annotated with a ground-truth oppression score by human experts, which serves as the reference label. Model outputs are compared against these annotations. To comprehensively assess performance, we use five metrics: (i) mean absolute error (MAE), (ii) classification accuracy (Acc), (iii) Cohen’s kappa ($\kappa$), (iv) Pearson correlation ($r$), and (v) Spearman correlation ($\rho$). 

\subsubsection{Baselines}\label{sec:baseline}
To demonstrate the contribution of the rule-guided prompt module, we compare against two prompting strategies, which is illustrated in Figure \ref{fig:promptchart}:
\begin{itemize}[leftmargin=*]
    \item \textbf{Vanilla prompting} – Directly asks the model to assign a score using only the oppression schema without additional reasoning instructions.
    \item \textbf{Chain-of-Thought (CoT) prompting} – Encourages the model to explain its reasoning step by step before providing a final score (appended after the oppression schema in the prompt):
\end{itemize}
\begin{quote}
    ``\emph{Think step by step before coming to your final answer. Start by answering these questions: } 
\begin{enumerate}
    \item \emph{Has this group historically faced institutional oppression in \texttt{\{country\}}?} 
    \item \emph{How were they positioned within dominant institutions (government, education, legal system) of \texttt{\{country\}}?} 
    \item \emph{Did they experience long-term structural disadvantage compared to dominant groups in \texttt{\{country\}}?} 
\end{enumerate}
\emph{After answering each question, decide which oppression scale level (1–5) fits best.}''
\end{quote}

\subsubsection{Language Models}\label{sec:language_models}
For generalizability, we evaluate these prompting strategies across three state-of-the-art LLMs:
\begin{itemize}[leftmargin=*]
    \item \gemini~\cite{team2024gemini}: A multimodal model developed by Google DeepMind.
    \item \gptiii~\cite{ouyang2022training}: An instruction-tuned model created by OpenAI.
    \item \gptivo~\cite{hurst2024gpt}: A fast, low-latency variant of GPT-4 released by OpenAI.
\end{itemize}

\subsubsection{Implementation Details}\label{sec:implementation_detail}
Each prompt is dynamically constructed using the LangChain framework and deployed via the respective model API in a zero-shot configuration, ensuring generalizability across global contexts. Prompts are executed independently for each identity--country pair, with Python’s \texttt{ThreadPoolExecutor} used for parallelization. All experiments are conducted with temperature set to 0 to enforce deterministic outputs.

This study was performed in accordance with the ethical standards established by the 1964 Declaration of Helsinki and its later amendments. The University of Rochester’s Research Subjects Review Board determined that this study met federal and University criteria for exemption (STUDY00004825).

\subsection{Main Results}\label{sec:result}

\begin{table}[ht]
\centering
\caption{Model performance under different prompting strategies. The best results for each model are \textbf{highlighted in bold}.}
\adjustbox{width=\linewidth}{
\begin{tabular}{c|l|ccccc}
\toprule
\textbf{LLM}& 
\multicolumn{1}{c|}{\textbf{Prompt}}& 
\textbf{MAE $\downarrow$}& 
\textbf{Acc $\uparrow$}& 
\textbf{$\kappa$ $\uparrow$} &
\textbf{$r$ $\uparrow$} &
\textbf{$\rho$ $\uparrow$} \\
\midrule
\multirow{3}{*}{\gemini}  & Vanilla &0.509 & 0.515 & 0.374 & 0.831 & 0.837\\
& CoT & 0.560 & 0.512 & 0.374 & 0.781 & 0.773\\
& Rule-guided & \textbf{0.401} & \textbf{0.608} & \textbf{0.482} & \textbf{0.852} & \textbf{0.844}\\
\hline
\multirow{3}{*}{\gptiii}  & Vanilla & 0.614 & 0.419 & 0.239 & 0.733 & 0.678\\
& CoT& 0.749 & 0.353 & 0.190 & 0.692 & 0.693 \\
& Rule-guided& \textbf{0.521} & \textbf{0.542} & \textbf{0.408} & \textbf{0.755} & \textbf{0.729}\\
\hline
\multirow{3}{*}{\gptivo}  & Vanilla   & 0.461 & \textbf{0.581} & \textbf{0.448} & 0.801 & 0.814 \\
&CoT    & 0.497 & 0.536 & 0.398 & 0.819 & \textbf{0.836} \\
&Rule-guided  & \textbf{0.446} & 0.575 & 0.440 & \textbf{0.820} & 0.831 \\
\bottomrule
\end{tabular}
}
\label{tab:full}
\end{table}

Across all models and prompting strategies (Table~\ref{tab:full}), {\gemini} achieved the best overall performance. Under the rule-guided setting, it produced the lowest mean absolute error (MAE = 0.401), the highest accuracy (0.608), and the strongest agreement with human expert annotations ($\kappa$ = 0.482; $r$ = 0.852; $\rho$ = 0.844). In contrast, {\gptiii} consistently underperformed, yielding higher MAE and weaker agreement across all metrics. {\gptivo} delivered competitive results—particularly under the Vanilla prompt (Acc = 0.581; $\kappa$ = 0.448)—but did not surpass {\gemini} in the rule-guided setting. These findings suggest that while {\gptivo} performs well in baseline classification, its smaller architecture and efficiency-oriented design may constrain its ability to capture nuanced, multi-layered sociological rules, making {\gemini} better suited for structured, theory-informed reasoning tasks.
% \newpage
\begin{figure}[h!]
    \centering
    \includegraphics[width=1\linewidth]{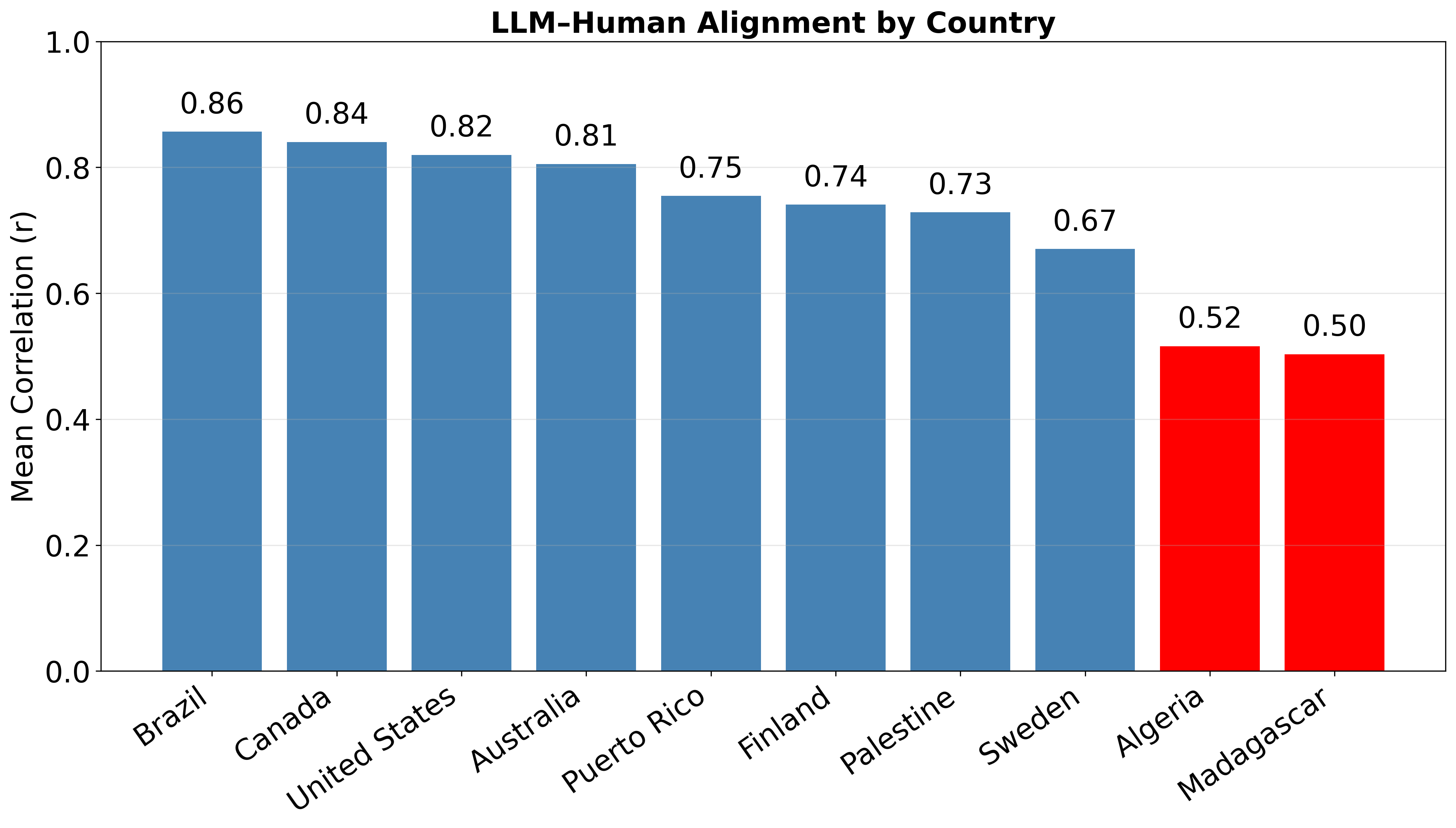}
    \caption{Pearson correlations between LLM predictions and human annotations across countries. Brazil has the highest alignment at $r$ = 0.86 while Algeria and Madagascar have the lowest alignment around $r$ = 0.5.}
    \label{fig:country_align}
\end{figure}

At the country level (Figure~\ref{fig:country_align}), the strongest alignment with human annotations was observed in Brazil ($r$ = 0.86), followed by Canada (0.84), the United States (0.82), and Australia (0.81). Moderate correlations appeared in Puerto Rico (0.75), Finland (0.74), Palestine (0.73), and Sweden (0.67). In contrast, alignment declined markedly in Algeria (0.52) and Madagascar (0.50). Specific cases highlight this discrepancy, as identity labels such as ``Arabo-musulman" and ``algérien" were frequently misclassified as ``high oppression", with the models attributing disadvantage to the legacy of colonization and systemic exclusion, even though these groups represent the national majority. In contrast, the lower correlation in Madagascar is more reflective of sample size than misclassification, as most identities differed a single level only with no major divergences. These results suggest that LLMs perform more reliably in Western countries with dominant global sociopolitical narratives, but may struggle in contexts requiring more domain-specific or historically grounded knowledge, or in settings with limited data. We provide a more detailed discussion in the following error analysis.

\subsection{Error Analysis}\label{sec:error_analysis}

Given the reasonable alignment performance across all the countries, we further analyze the distribution of errors in this section. 
\begin{figure}[h!]
    \centering
    \includegraphics[width=1\linewidth]{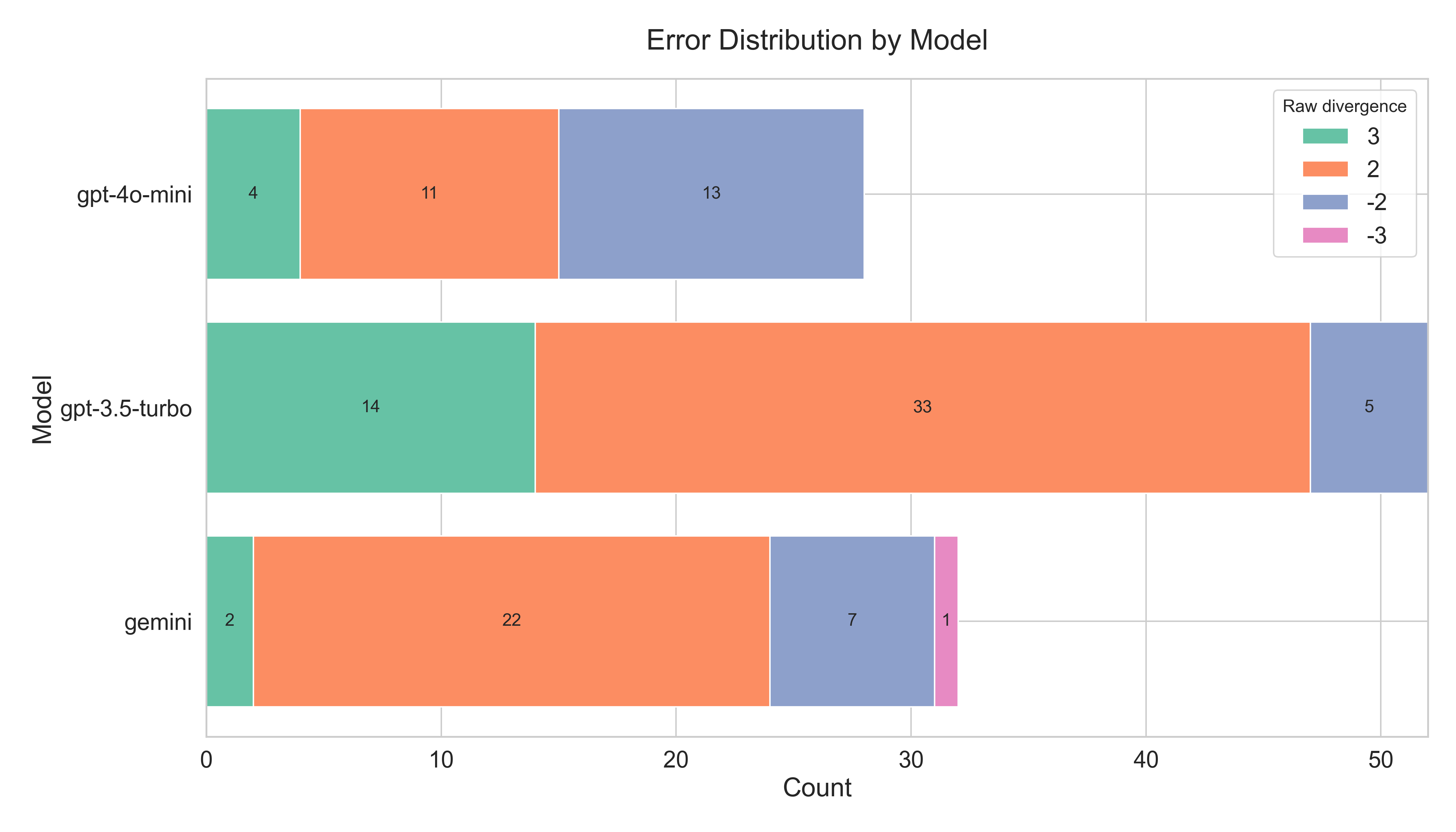}
    \caption{Distribution of absolute differences between LLM-predicted oppression scores and human expert annotations across 334 identity-country pairs, grouped by model. GPT-3.5-Turbo produces the highest number of high-magnitude errors ($|\text{difference}| \geq 2$), while GPT-4o-mini produces the fewest deviations.}
    \label{fig:error_analysis_model}
\end{figure}

\subsubsection{Incorrect Estimation of Severity}
Overestimation of severity occurs when a model assigns a higher level of oppression to a response compared to the assigned score provided by human annotators. In contrast, underestimation of severity occurs when a model assigns a lower level of oppression to a response compared to the assigned response provided by human annotators.

If excluding assignments with absolute difference of 1 (where model assignments have one level over or one level under human assignments), across models, {\gptiii} shows the biggest pattern of overestimation, with 14 responses having three levels over and 33 responses having two levels over the human assignments, followed by {\gemini} with two responses having three levels over and 22 responses having two levels over. Meanwhile, {\gptivo} has the highest number of underestimation of severity, with 13 responses being assigned two levels under the human assignments (Figure~\ref {fig:error_analysis_model}). In total, all models show a behavior of more overestimation of severity than underestimation. 

\begin{figure}[h!]
    \centering
    \includegraphics[width=1\linewidth]{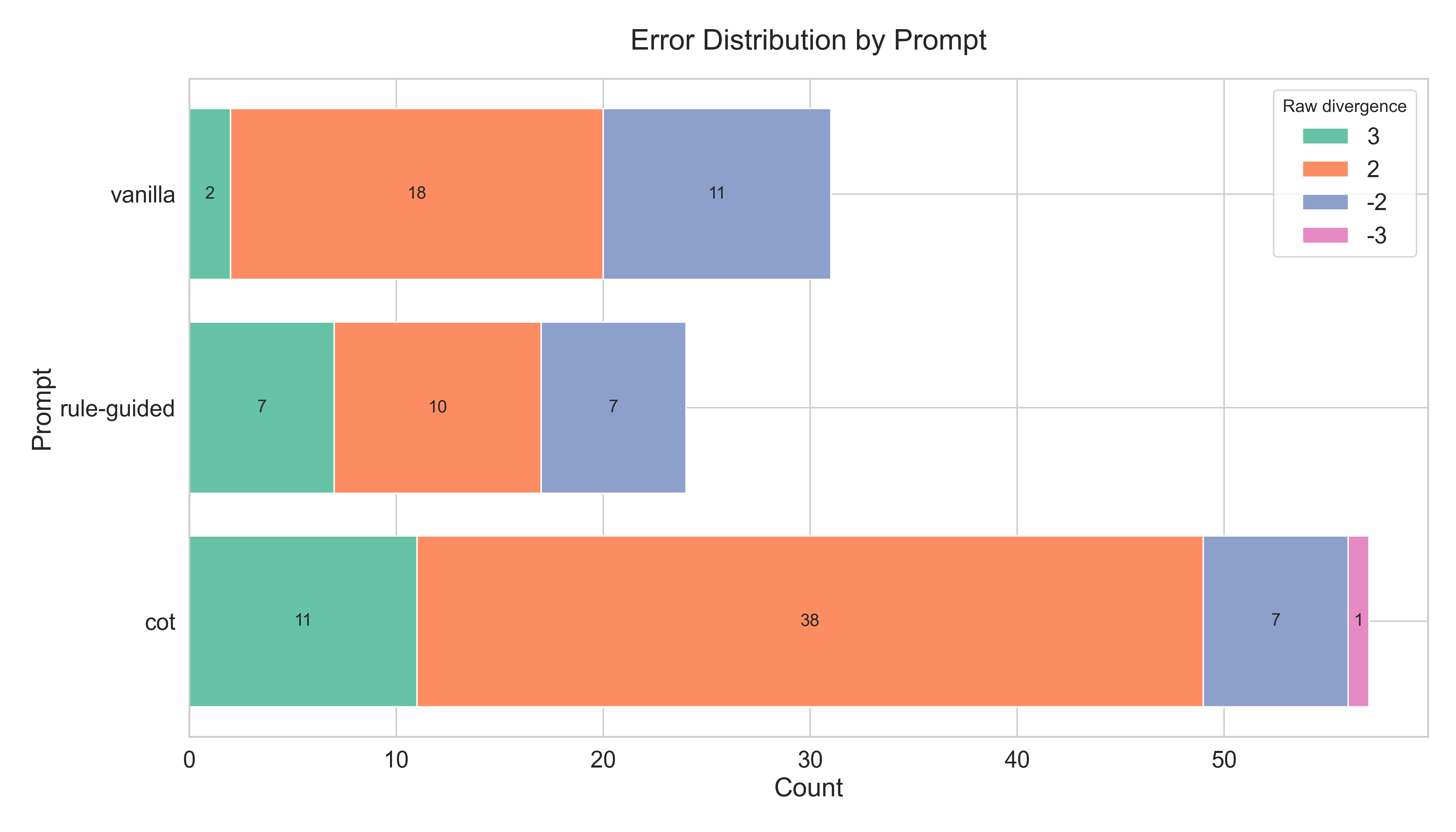}
    \caption{Distribution of absolute differences between LLM-predicted oppression scores and human expert annotations across 334 identity-country pairs, grouped by prompt method. Chain-of-Thought (CoT) prompting produces the highest number of high-magnitude errors ($|\text{difference}| \geq 2$), while Rule-Guided prompting produces the fewest deviations.}
    \label{fig:error_analysis_prompt}
\end{figure}

Similarly, across all prompts (Figure~\ref{fig:error_analysis_prompt}), Chain-of-Thought (CoT) prompting shows the biggest pattern of overestimation, with 11 responses having three levels over human estimation, 38 responses two levels over, 7 responses two levels below, and one response three levels under human estimation. Rule-guided prompts exhibit the least number of errors, with only 7 errors at three levels above human annotations, 10 responses at two levels above, and 7 responses at two levels below.

\subsubsection{Reasons for Incorrect Estimations} 
The models often provide one of these three analysis patterns for the incorrect assignments: i) misunderstanding the provided response or wrong assumption about the provided response with regard to the specific country related to that response, ii) over or under-focusing on the instances of oppression faced by the identified group in the response, and iii) hallucination. 

\begin{figure}[h!]
    \centering
    \includegraphics[width=1\linewidth]{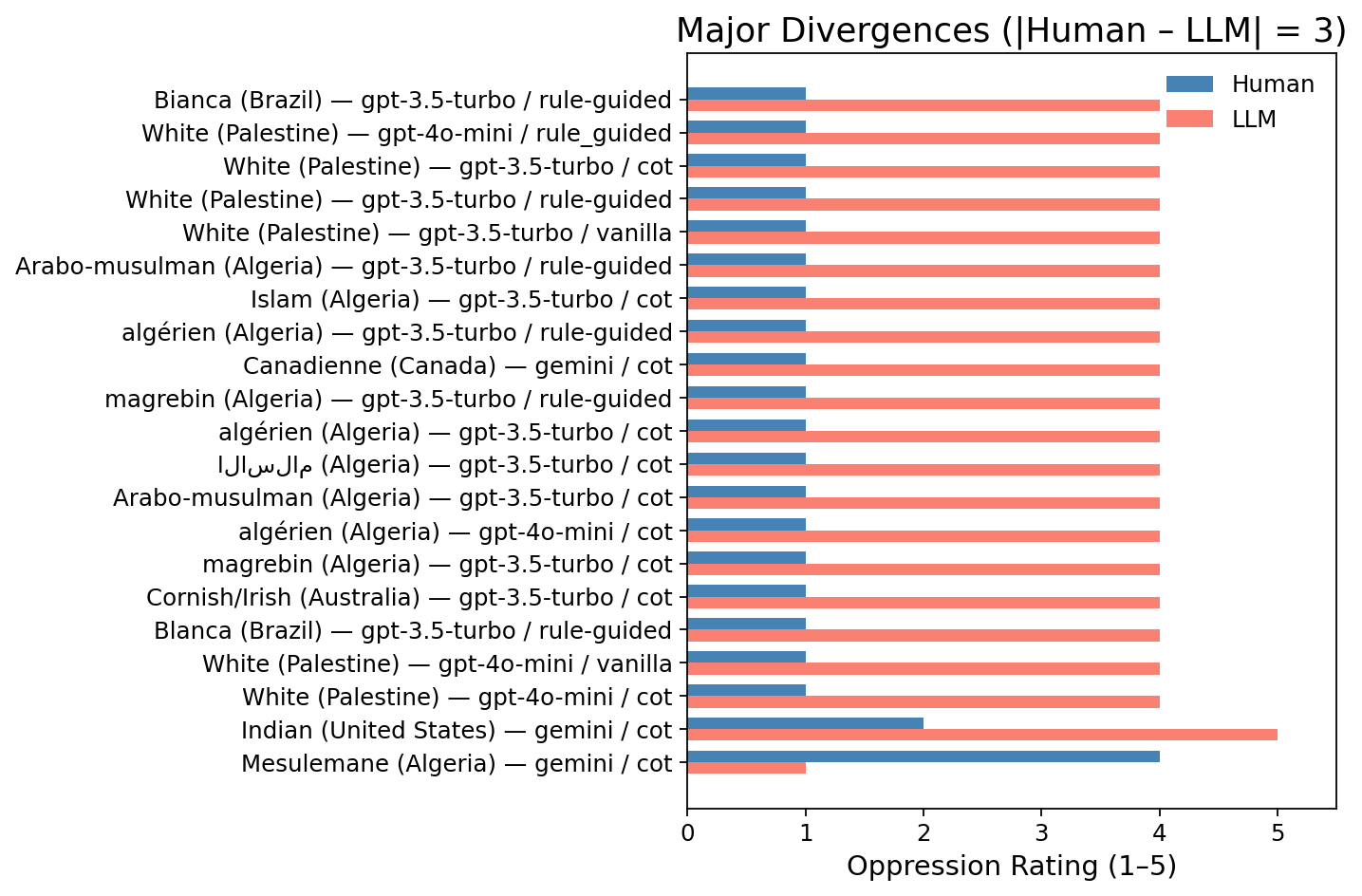}
    \caption{Highest Divergence between Human and LLM ratings across all models and prompt strategies. Majority of divergences occur from identities in Algeria, or are produced from scores generated by either CoT prompting or GPT-3.5-turbo.}
    \label{fig:divergent_cases}
\end{figure}
Among all of the top divergences in our experiments, the majority are either produced by {\gptiii}, in the CoT setting, or are identities in the country of Algeria (Figure~\ref{fig:divergent_cases}). One example of CoT prompting leading to misunderstanding in the provided response is shown in this assignment provided by {\gemini}: For response ``Indian" in country ``United States", the human annotator assigned an oppression level of 2, with the assumption that the response refer to the identity of \textit{Americans of Indian origin}. The model assigned this response a level of 5, with the assumption that the response refers to \textit{Native American}, arguing that \textit{``... [Native Americans/Indigenous peoples in the United States] have experienced extreme forms of violence, exclusion, and marginalization, resulting in long-term structural disadvantage."} 
While the reasoning itself is valid, the divergence arises from the model's incorrect assumption about which identity the response denotes.

Another top divergence occurs with the response ``White" in ``Palestine", which is consistently assigned a score of 4 (High Oppression) rather than 1 (Little to No Oppression). In this scenario, the models appear to conflate the broader systemic oppression of Palestinians with the specific identity label provided, producing justifications that emphasize colonization and legal discrimination while overlooking the privileged status of ``White" in that context. This dynamic is similarly spotted in responses such as ``Cornish/Irish" in ``Australia" and ``Blanca" in ``Brazil", where {\gptiii} projects histories of systemic exclusion that the annotators do not apply in their ranking.

Similarly, divergences are also common for identities from Algeria, such as ``Arabo-musulman", ``magrebin", and ``algérien". These identities are consistently scored as 4s compared to their ground-truth label of 1. Conversely, identities such as ``North African" and ``Nord-africaine" are scored as a 2 by {\gptivo}, and ``Musulemane" is scored as 1 by {\gemini} when human annotators assigned these these identities a score of 4. These results suggest that the models struggle with distinguishing between historically marginalized groups and majority populations in Algeria.

Overall, these examples of divergence are most pronounced within identity terms that are more ambiguous in nature or politically contested, causing the models to overgeneralize broad narratives of oppression.

\section{Discussion}\label{sec:discussion}
Our results reveal that Large Language Models (LLMs) can approximate judgments of historical identity-based oppression from free-text self-reported ethnicity and country inputs, but only when guided by structured, theory-informed prompts. Across all models, the rule-guided approach consistently reduces error and improves alignment with expert annotations relative to vanilla and Chain-of-Though (CoT) baselines. {\gemini} achieves the strongest overall performance in this setting (MAE = 0.401; Acc = 0.608; $\kappa$ = 0.482; $r$ = 0.852; $\rho$ = 0.844). These results underscore that unguided CoT reasoning does not reliably enhance performance without domain-specific rules. By explicitly encoding how ambiguous cases should be handled, what counts as evidence, and what higher levels should be assigned, the rule-guided rubric constrains model reasoning and promotes more historically grounded, context-sensitive judgments.

Additionally, LLMs align more closely with human annotations in Brazil, Canada, the United States, and Australia, while showing weaker correlation in Madagascar and Algeria, and mixed outcomes in Puerto Rico and Palestine. This suggests that the models perform better in regions with well-documented historical narratives, and struggle in regions where histories of exclusion are less represented in training data or are more politically complex.

All models display a bias of overestimation, with {\gptiii} and CoT prompting each exhibiting the greatest number of errors among models and prompting styles, respectively. In contrast, {\gptivo} showed the least number of errors among models, and the rule-guided approach yields the fewest among prompt types. Most errors are clustered around (i) misinterpreting ambiguous identities (\eg mapping ``Indian" in the United States to ``Indigenous" instead of South Asian), (ii) overweighting or underweighting specific discrimination patterns, and (iii) hallucinating historical details. These patterns point to built-in model biases that flatten nuanced distinctions, highlighting the need for explicit rules to handle ambiguity and set consistent thresholds.

This study has limitations. Our current study only examines ethnicity-related identity, not other important aspects of identity (e.g., gender identity, sexual orientation, disability identity). Additionally, the annotated set (n=334) spans 10 countries and, while intentionally diverse, cannot represent the full global distribution of identities and contexts. Similarly, the sample sizes for certain countries are relatively small. While we report point estimates of model performance (e.g., Pearson’s $r$, Spearman’s $\rho$, Cohen’s $\kappa$), we do not include 95\% confidence intervals around these statistics, which would more fully capture the uncertainty introduced by small group sizes. As such, these metrics should be interpreted as suggestive rather than definitive; future work with larger, more balanced datasets is needed to validate the observed patterns and assess the consistency of model performance across diverse subpopulations. The transformation of multidimensional oppression into a five-level ordinal scale necessarily abstracts complex realities; different rubrics or task decompositions (\eg legal status, economic exclusion, exposure to violence) might yield different performance profiles. Although we train annotators, adjudicate disagreements, and document sources and rationales, expert judgment is itself contextual and may reflect scholarly debates or data gaps in particular regions or languages. We also did not compute intercoder agreement metrics such as Cohen's $\kappa$ for the ground-truth annotations and instead used a consensus-based approach. This method should be complemented with formal reliability assessments in future work to strengthen reproducibility. Finally, we conduct all prompting in a standardized template; alternative designs (\eg richer exemplars, multi-turn retrieval-augmented reasoning) could shift outcomes.

Overall, our findings highlight the importance of structured, theory-informed prompting in aligning LLM outputs with expert-annotated oppression scores. Reasoning under rubric-guided rules, rather than unguided Chain-of-Thought (CoT) reasoning, is most effective at reducing error and improving reliability. Despite these additions, systemic miscalibration exists, particularly for ambiguous identities and politically complex regions, reinforcing the need for calibration against human-annotated benchmarks. Future work could incorporate Retrieval-Augmented Generation (RAG) or few-shot calibration to reduce bias in underperforming regions. Integrating country-specific histories via dynamic context windows could further enhance cultural specificity of LLM judgements.

\section{Conclusion}\label{sec:conclusion}
This work introduces a novel approach to measuring oppression by constructing a bottom-up schema from self-reported free-text ethnicity and residence information and scaling the measurement through LLMs. Our framework complements existing indices of structural oppression by foregrounding identity-based exclusion and lived experience. Through rule-guided prompting, we enhance the validity and interpretability of model outputs, achieving strong alignment with expert annotations (Pearson $r$ = 0.852) and establishing the first benchmark on identity-based oppression classification.

Taken together, our findings show that LLMs, when carefully guided, can serve as scalable tools for capturing historical oppression as a specific form of systemic social disadvantage. At the same time, the identified limitations point to the need for continued methodological development to mitigate biases and ensure robustness across diverse cultural and geopolitical contexts. By releasing both the dataset and benchmark results, we provide a foundation for future research to build more accurate, fair, and context-sensitive approaches to understanding oppression in data-driven social science.

\section*{Acknowledgments}
We used ChatGPT for language editing to improve clarity and assist with LaTeX formatting. The use of generative AI in this manuscript adheres to ethical guidelines for use and acknowledgment of generative AI in academic research, as outlined in Mann \textit{et al.}~\cite{porsdam2024guidelines}. This manuscript has been thoroughly vetted for accuracy, and we assume all responsibility for the integrity of this contribution.

\bibliographystyle{IEEEtran}
\bibliography{custom}

\appendix

\subsection{Additional Details of Methods}\label{appendix_sec:method}
\subsubsection{Survey Question}\label{appendix_sec:survey_question}

%\textit{How would you describe your ethnic background?}

We provide the question in several other languages below:

\begin{itemize}
\item \textbf{Italian:} \textit{Come descriveresti le tue origini etniche?}
\item \textbf{French:} \textit{Comment décririez-vous vos origines ethniques?}
\item \textbf{Spanish:} \textit{¿Cómo describiría su trasfondo étnico?}
\item \textbf{Mandarin:} \includegraphics[trim=10pt 14pt 10pt 9pt, clip=true, height=0.8em]{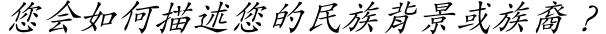}
\item \textbf{Arabic:} \includegraphics[trim=10pt 14pt 10pt 9pt, clip=true, height=0.8em]{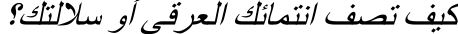}
\item \textbf{Hindi:} \includegraphics[trim=10pt 14pt 10pt 9pt, clip=true, height=0.8em]{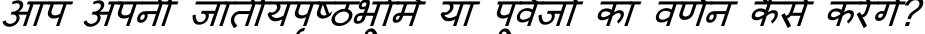}
\item \textbf{American Sign Language (ASL):} \url{https://youtu.be/p8AJYSlubkM}
\end{itemize}

\end{document}